\documentclass[10pt,twocolumn,letterpaper]{article}

\usepackage{cvpr}
\usepackage{times}
\usepackage{epsfig}
\usepackage{graphicx}
\usepackage{amsmath}
\usepackage{amssymb}
\usepackage{float}
\usepackage{booktabs}       

\usepackage[pagebackref=true,breaklinks=true,letterpaper=true,colorlinks,bookmarks=false]{hyperref}

\cvprfinalcopy 

\def\cvprPaperID{446} 

\ifcvprfinal\fi
\begin{document}

\title{Perceptual Generative Adversarial Networks for Small Object Detection}

\author{\normalsize{Jianan~Li \quad Xiaodan~Liang \quad Yunchao~Wei \quad Tingfa~Xu \quad Jiashi~Feng \quad Shuicheng~Yan}\\
	{} {}
}

\maketitle
\thispagestyle{empty}

\begin{abstract}
Detecting small objects is notoriously  challenging due to their low resolution and noisy  representation. Existing object detection pipelines usually detect small objects through learning representations of all the objects at multiple scales. However, the performance gain of such ad hoc architectures is usually limited to pay off the computational cost.
In this work, we address the small object detection problem by developing a single  architecture that internally lifts representations of small objects to  ``super-resolved'' ones,  achieving similar characteristics as large objects and thus more discriminative for detection. For this purpose, we propose a new Perceptual Generative Adversarial Network (Perceptual GAN) model that  improves small object detection  through  narrowing representation difference of small  objects from the large ones. Specifically, its generator learns to  transfer perceived   poor representations of the small objects to super-resolved ones  that are  similar enough to real large objects to fool a competing discriminator. Meanwhile its discriminator  competes  with the generator to identify the generated  representation and imposes an additional perceptual requirement~\textendash~generated representations of small objects must be beneficial for  detection purpose~\textendash~on the generator. Extensive evaluations on the challenging Tsinghua-Tencent 100K~\cite{zhu2016traffic} and the Caltech~\cite{dollar2012pedestrian} benchmark well demonstrate the superiority of Perceptual GAN in detecting small objects, including traffic signs and pedestrians, over well-established state-of-the-arts.

\end{abstract}
\vspace{-1em}

\section{Introduction}
\vspace{-1mm}

\begin{figure}[h]
	\centering
	\includegraphics[scale=0.45]{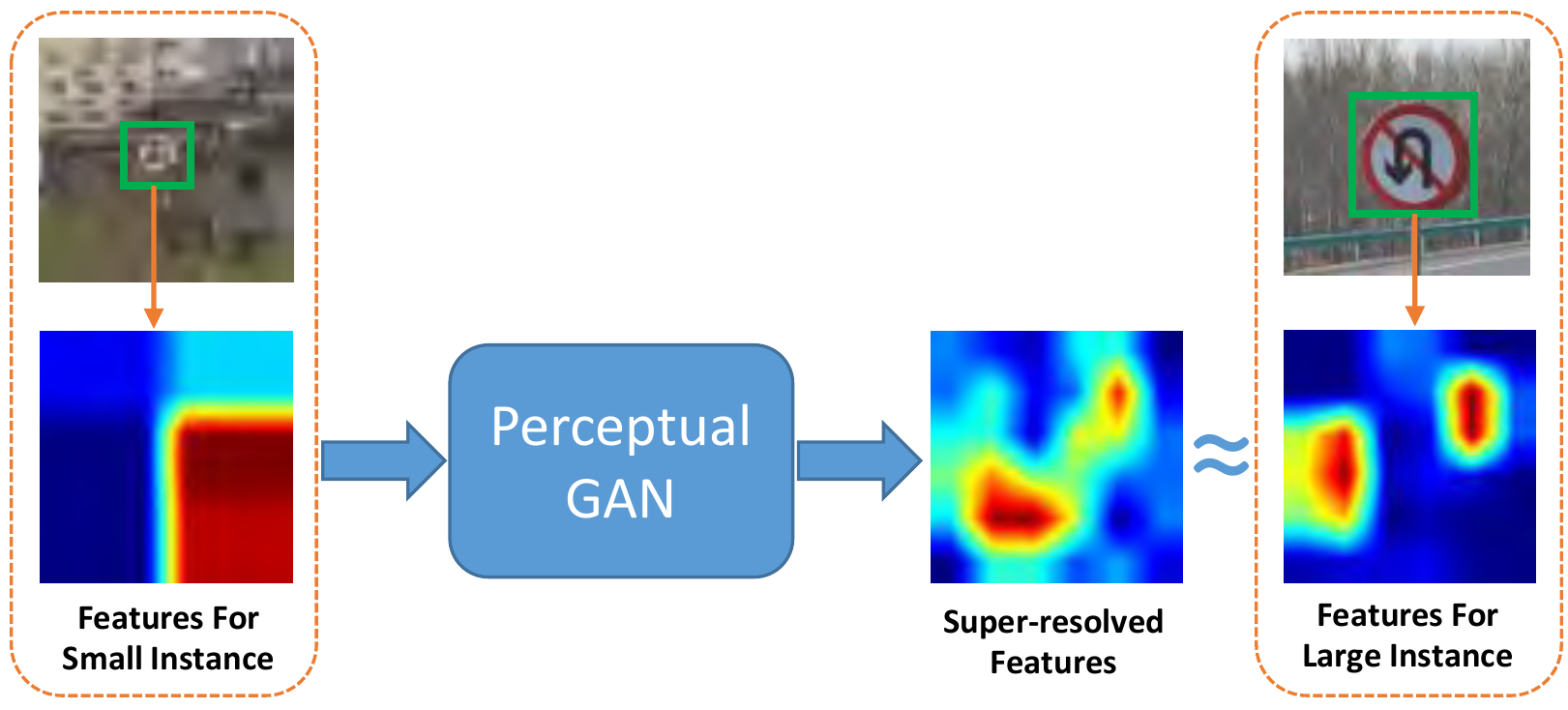}
	\caption{Large  and small objects exhibit different  representations from high-level convolutional layers of a CNN detector. The representations of large objects are discriminative while those of small objects are of low resolution, which hurts the detection accuracy.  In this work, we introduce the Perceptual GAN model to enhance the  representations for small objects to be similar to real large objects, thus improve detection performance on the small objects.}
	\vspace{-4mm}
	\label{fig:motivation}
	\vspace{-2mm}
\end{figure}

Recent great progress on object detection is stimulated by the deep learning pipelines that learn deep representations from the region of interest (RoI) and perform classification based on the learned representations, such as Fast R-CNN~\cite{girshick2015fast} and Faster R-CNN~\cite{ren2015faster}. Those pipelines indeed work well on large objects with high resolution, clear appearance and structure from which the discriminative features can be learned. But they usually fail to detect \emph{very small} objects, as rich representations are difficult to learn from their poor-quality appearance and structure, as shown in Figure~\ref{fig:motivation}. However, small objects are very common in many real world applications such as traffic sign detection, pedestrian detection for advanced autonomous driving. Small object detection is much more challenging than normal object detection and good solutions are still rare so far.

Some efforts~\cite{chen20153d,liu2015ssd,jiang2016object,yang2016exploit,li2015convolutional,bell2015inside} have been devoted to addressing small object detection problems. One common practice~\cite{chen20153d,liu2015ssd} is to increase the scale of input images to enhance the resolution of small objects and produce high-resolution feature maps. Some others~\cite{yang2016exploit,li2015convolutional,bell2015inside} focus on developing network variants to generate multi-scale  representation which enhances high-level small-scale features with multiple lower-level features layers. 
However, all of those approaches try to enhance the performance of small object detection by data augmentation or naively increasing the feature dimension. Simply increasing the scale of input images often results in heavy time consumption for training and testing. Besides, the multi-scale representation constructed by the low-level features just works like a black-box and cannot guarantee the constructed features are interpretable and discriminative enough for object detection. In this work, we argue that a preferable way to effectively represent the small objects is to discover the intrinsic structural correlations between small-scale and large-scale objects for each category and then use the transformed representations to improve the network capability in a more intelligent way.

Therefore, we propose a novel Perceptual Generative Adversarial Network (Perceptual GAN) to generate super-resolved  representations for small objects for better detection. The Perceptual GAN aims to enhance the representations of small objects to be similar to those of large object, through fully exploiting the structural correlations between objects of different scales during the network learning. It consists of two subnetworks, \textit{i.e.}, a generator network and a perceptual discriminator network. Specifically, the generator is a deep residual based feature generative model which transforms the original poor features of small objects to highly discriminative ones by introducing fine-grained details from lower-level layers, achieving ``super-resolution" on the intermediate representations. The discriminator network serves as a supervisor and provides guidance on the quality and advantages of the generated fine-grained details. Different from the vanilla GAN, where the discriminator is only trained to differentiate fake and real representations, our proposed Perceptual GAN includes a new perceptual loss tailored for the detection purpose. Namely, the discriminator network is trained not only to differentiate between the generated super-resolved  representations for small objects and the original ones from real large objects with an adversarial loss, but also to justify the detection accuracy benefiting from the generated super-resolved features with a perceptual loss. 

We optimize the parameters of the generator and the discriminator network in an alternative manner to solve the min-max problem. In particular, the generator network is trained with the goal of fooling the discriminator by generating the most large-object like representations from small objects as well as benefiting the detection accuracy. On the other hand, the discriminator is trained to improve its discriminative capability to correctly distinguish the generated super-resolved  representations from those from real large objects, and also provides feedback about the localization precision to the generator. Through competition between these two networks, generator is effectively trained to enhance the  representations for small objects to super-resolved ones capable of providing high detection accuracy.


We evaluate our Perceptual GAN method on the challenging Tsinghua-Tencent 100K~\cite{zhu2016traffic} and the Caltech benchmark~\cite{dollar2012pedestrian} for traffic sign and pedestrian detection respectively.  Small instances are common on these two datasets, thus they provide suitable testbed for evaluating methods on detecting small objects. Our proposed method shows large improvement over  state-of-the-art methods and demonstrates its superiority on detecting small objects.

To sum up, this work makes the following contributions. (1) We are the first to successfully apply GAN-alike models to solve the challenging small-scale object detection problems. (2) We introduce a new conditional generator model that learns the additive residual representation between large and small objects, instead of generating the complete  representations as before. (3) We introduce a new perceptual discriminator that provides more comprehensive supervision beneficial for detections, instead of barely differentiating fake and real. (4) Successful applications on traffic sign detection and pedestrian detection have been achieved with the state-of-the-art performance. 

\section{Related Work}

\subsection{Small Object Detection}
\paragraph{Traffic Sign Detection}
Traffic sign detection and recognition has been a popular problem in intelligent vehicles, and various methods~\cite{le2010real,haloi2015novel,sermanet2011traffic,jin2014traffic,wu2013traffic,zhu2016traffic} have been proposed to address this challenging task. Traditional  methods for this task includes~\cite{le2010real}~\cite{haloi2015novel}. Recently, CNN-based approaches have been widely adopted in traffic sign detection and classification due to their high accuracy. In particular, Sermanet \emph{et al.}~\cite{sermanet2011traffic} proposed to feed multi-stage features to the classifier using connections that skip layers to boost traffic sign recognition. Jin \emph{et al.}~\cite{jin2014traffic} proposed to train the CNN with hingle loss, which provides better test accuracy and faster stable convergence. Wu \emph{et al.}~\cite{wu2013traffic} used a CNN combined with fixed and learnable filters to detect traffic signs. Zhu \emph{et al.}~\cite{zhu2016traffic} trained two CNNs for simultaneously localizing and classifying traffic signs.

\vspace{-3mm}
\paragraph{Pedestrian Detection}
The hand-crafted features achieve great success in pedestrian detection. For example, Doll{\'a}r \emph{et al.} proposed Integral Channel Features (ICF)~\cite{dollar2009integral} and Aggregated Channel Features (ACF)~\cite{dollar2014fast}, which are among the most popular hand-crafted features for constructing pedestrian detectors. Recently, deep learning methods have greatly boosted the performance of pedestrian detection~\cite{ouyang2013joint,sermanet2013pedestrian,ouyang2012discriminative,ta_cnn,zhang2016faster}. Ouyang \emph{et al.}~\cite{ouyang2013joint} proposed a deformation hidden layer for CNN to model mixture poses information, which can further benefit the pedestrian detection task. Tian \textit{et al.}~\cite{ta_cnn} jointly optimized the pedestrian detection with semantic tasks. Sermanet \emph{et al.}~\cite{sermanet2013pedestrian} utilized multi-stage features to integrate global shape information with local distinctive information to learn the detectors.

\vspace{-2mm}
\subsection{Generative Adversarial Networks}
\vspace{-2mm}
The Generative Adversarial Networks (GANs)~\cite{goodfellow2014generative} is a framework for learning generative models. Mathieu \emph{et al.}~\cite{mathieu2015deep} and Denton\emph{et al.}~\cite{denton2015deep} adopted GANs for the application of image generation. In~\cite{li2016combining} and~\cite{yeh2016semantic}, GANs were employed to learn a mapping from one manifold to another for style transfer and inpainting, respectively. The idea of using GANs for unsupervised representation learning was described in  \cite{radford2015unsupervised}. GANs were also applied  to image super-resolution in \cite{ledig2016photo}.  To the best of our knowledge, this work makes the first attempt to accommodate GANs on the object detection task to address the small-scale problem by generating super-resolved  representations for small objects. 

\begin{figure}[t]
	\centering
	\includegraphics[width=0.48\textwidth]{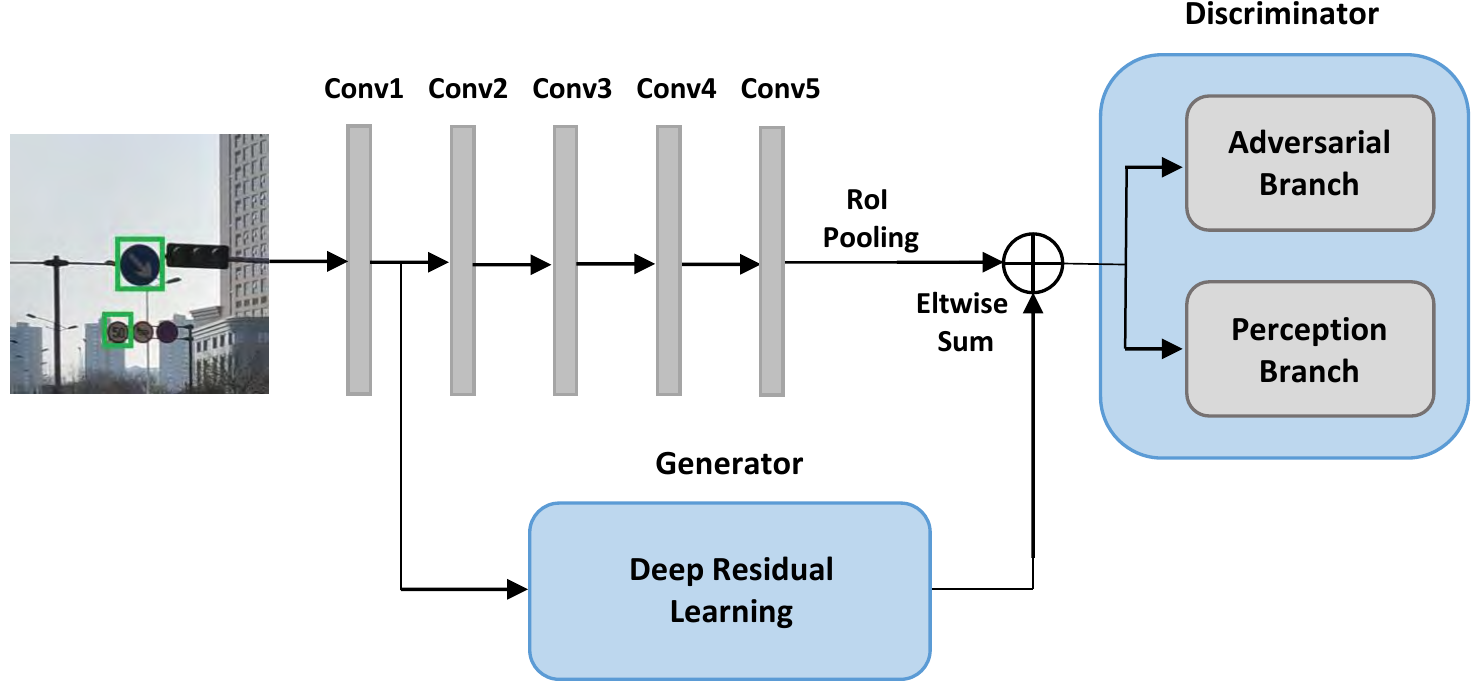}
	\caption{Training procedure of object detection network based on the Perceptual GAN. The perception branch of the discriminator network along with the bottom convolutional layers is first trained using the training images that contain only large objects. Then with the training images that contain only small objects, the generator network is trained to generate super-resolved large-object like  representations for small objects. The adversarial branch of the discriminator network is trained to differentiate between the generated super-resolved  representations for small objects and the original ones for real large objects. By iteratively boosting the abilities of the generator network and the discriminator network through alternative training, the detection accuracy especially for small objects can be improved.}
	\label{fig:framework}
	\vspace{-1.5em}
\end{figure}

\vspace{-0.5em}
\section{Perceptual GANs}
\vspace{-0.5em}
We propose a new Perceptual GAN network to address the challenging small object detection problems. We introduce new designs on the generator model that is able to generate super-resolved  representations for small objects, and also a new discriminator considering adversarial loss and perceptual loss to ``supervise'' the generative process. In this section, we first present the alternative optimization for perceptual GAN from a global view. Then, the details of the generator for super-resolved feature generation and the discriminator for adversarial learning are given.

\begin{figure*}[t]
	\centering
	\includegraphics[scale=0.5]{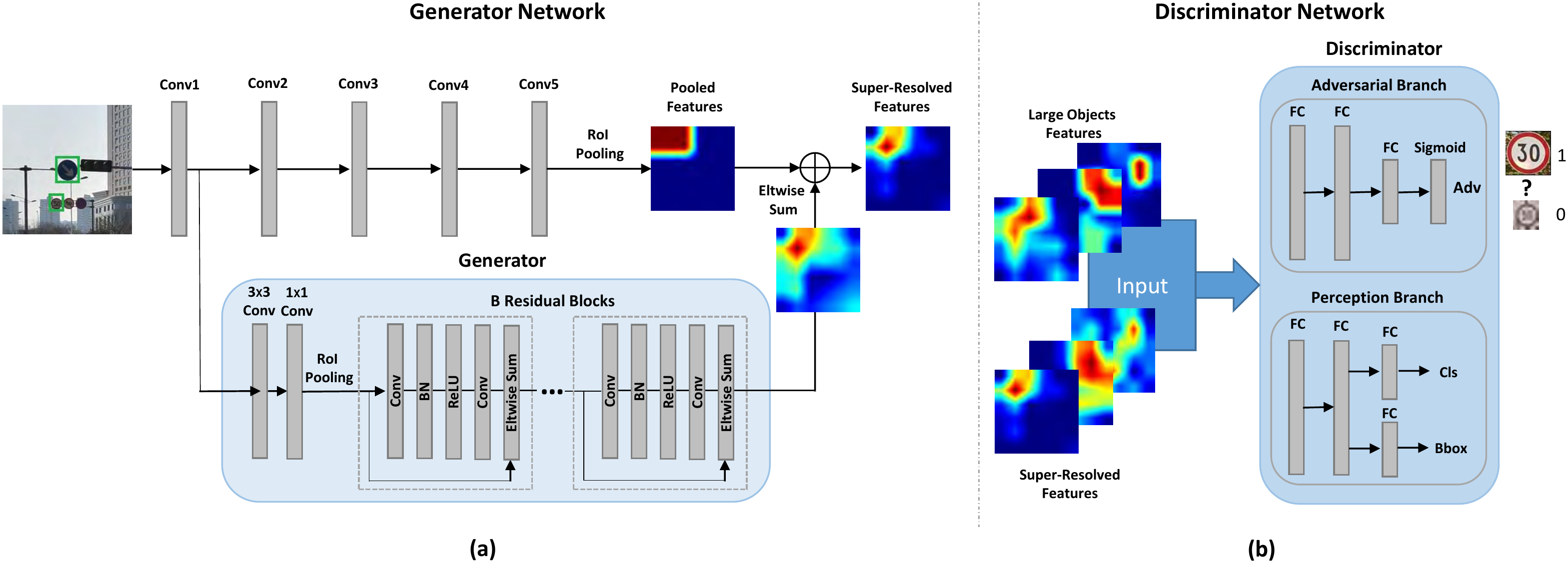}
	\caption{Details of the proposed Perceptual Generative Adversarial network. (a) The generator is a deep residual network which takes the features with fine-grained details from lower-level layer as input and passes them to $3\times3$ convolutional filters followed by $1\times1$ convolutional filters to increase the feature dimension to be aligned with that of ``Conv5". Then $B$ residual blocks each of which consists of convolutional layers followed by batch normalization and ReLU activation are employed to learn the residual representation, which is used to enhance the pooled features from ``Conv5" for small objects to super-resolved representation through element-wise sum operation. (b) The discriminator takes the features of large object and the super-resolved representation of small object as inputs and splits into two branches. The adversarial branch consists of three fully connected layers followed by sigmoid activation, which is used to estimate the probability that the current input representation belongs to that of real large object. The perception branch consists of two fully connected layers followed by two output sibling layers, which are used for classification and bounding box regression respectively to justify the detection accuracy benefiting from the generated super-resolved representation. 
 }
	\label{fig:subnetwork}
	\vspace{-1.6em}
\end{figure*}

\vspace{-2mm}
\subsection{Overview}
\vspace{-2mm}

The learning objective for vanilla GAN models~\cite{goodfellow2014generative} corresponds to a minimax two-player game, which is formulated as
\vspace{-2mm}
\begin{equation*}
	\vspace{-2mm}
	\begin{split}
		\min_G \max_D L(D,G) \triangleq ~&\mathbb{E}_{x \sim p_{\text{data}(x)}} \log D(x) \\
		&+ \mathbb{E}_{z \sim p_z(z)}\left[\log(1-D(G(z)))\right],
	\end{split}
	\vspace{-1em}
\end{equation*}
where $G$ represents a generator that learns to map data $z$ from the noise distribution $p_z(z)$ to the distribution $p_{\text{data}(x)}$ over data $x$, and $D$ represents a discriminator that estimates the probability of a sample coming from the data distribution $p_{\text{data}(x)}$ rather than $G$. The training procedure for $G$ is to maximize the probability of $D$ making a mistake. 

In our case, $x$ and $z$ are the  representations for large objects and small objects, \emph{i.e.}, $F_l$ and $F_s$ respectively. We aim to learn a generator function $G$ that transforms the  representations of a small object $F_s$ to a super-resolved one $G(F_s)$ that is similar to the original one of the large object $F_l$. Learning the representation $G(F_s)$ for small objects matching the distribution of large object feature $F_l$ may be difficult due to the limited information contained in $F_s$. We thus introduce a new conditional generator model which is conditioned on the extra auxiliary information, \emph{i.e.}, the low-level features of the small object $f$ from which the generator learns to generate the residual representation between the  representations of large and small objects through residual learning instead.
\vspace{-2.5mm}
\begin{equation*}
	\vspace{-1em}
	\begin{split}
		\min_G \max_D &~L(D,G) \triangleq \mathbb{E}_{F_l \sim p_{\text{data}(F_l)}} \log D(F_l)\\
		 & + \mathbb{E}_{F_s \sim p_{F_s}(F_s|f)}[\log(1-D(\underbrace{F_s+G(F_s|f)}_{
			\text{residual learning}}))].
	\end{split}
	\vspace{-1em}
\end{equation*}
In this case, the generator training can be substantially simplified over directly learning the super-resolved  representations for small objects. For example, if the input representation is from a large object, the generator only needs to learn a zero-mapping. Besides, we introduce a perceptual loss on the discriminator to benefit the detection task as detailed below. 

As shown in Figure~\ref{fig:framework},  the generator network aims to generate super-resolved  representation for the small object. The discriminator includes two branches, \ie the adversarial branch for differentiating between the generated super-resolved  representation and the original one for the large object and the perception branch for justifying the detection accuracy benefiting from the generated  representation. We optimize the parameters embedded in the generator and the discriminator network in an alternative manner to solve the adversarial min-max problem.

Denote $G_{\Theta_{g}}$ as the generator network with parameters $\Theta_{g}$. We obtain $\Theta_{g}$ by optimizing the loss function $L_{dis}$
\vspace{-2mm}
\begin{equation}
	\begin{aligned}
		\Theta_{g} = \arg \min_{\Theta_{g}}L_{dis}(G_{\Theta_{g}}(F_s)),
	\end{aligned}
	\vspace{-0.8em}
\end{equation} 
where $L_{dis}$ is the weighted combination of the adversarial loss $L_{dis\_a}$ and the perceptual loss $L_{dis\_p}$ produced by the discriminator network, which is detailed in Section \ref{sec:g}.
We train the adversarial branch of the discriminator network to maximize the probability by assigning the correct label to both the generated super-resolved feature for the small object $G_{\Theta_{g}}(F_s)$ and the feature for the large object $F_l$. 

Suppose $D_{\Theta_{{a}}}$ is the adversarial branch of the discriminator network parameterized by $\Theta_{{a}}$. We obtain $\Theta_{{a}}$ by optimizing a specific loss function $L_a$:
\vspace{-2mm}
\begin{equation}
\vspace{-0.3em}
	\begin{aligned}
		\Theta_{a} = \arg \min_{\Theta_{a}}L_a(G_{\Theta_{g}}(F_s), F_l),
	\end{aligned}
	\vspace{-0.5em}
\end{equation} 
where the loss $L_a$ is defined as
\vspace{-2mm}
\begin{equation}
\vspace{-0.7em}
	\begin{aligned}
		L_a = -\log D_{\Theta_{a}}(F_l) -\log (1-D_{\Theta_{a}}(G_{\Theta_{g}}(F_s))).
	\end{aligned}
\end{equation} 
Eventually, $L_a$ encourages the discriminator network to distinguish the difference between the currently generated super-resolved  representation for the small object and the original one from the real large object.

To justify the detection accuracy benefiting from the generated super-resolved  representation, the perception branch should be first well trained based on the features of large objects to achieve high detection accuracy. Denote $D_{\Theta_{{p}}}$ as the perception branch of the discriminator network parameterized by $\Theta_{{p}}$. We obtain $\Theta_{{p}}$ by optimizing a specific loss function $L_{dis\_p}$ with the  representation for the large object: 
\vspace{-2mm}
\begin{equation}
\vspace{-0.5em}
	\begin{aligned}
		\Theta_{p} = \arg \min_{\Theta_{p}}L_{dis\_p}(F_l),
	\end{aligned}	
\end{equation} 
where $L_{dis\_p}$ is the multi-task loss for  classification and  bounding-box regression, which is detailed in Section \ref{sec:g}.

With the average size of all instances, we obtain two subsets containing small objects and large objects, respectively. For overall training, we first learn the parameters of bottom convolutional layers and the perception branch of the discriminator network based on the subset containing large objects. Guided by the learned perceptual branch, we further train the generator network based on the subset containing small objects and the adversarial branch of the discriminator network using both subsets. We alternatively perform the training procedures of the generator and the adversarial branch of the discriminator network until a balance point is finally achieved, \ie large-object like super-resolved features can be generated for the small objects with high detection accuracy. 


\vspace{-1.0mm}
\subsection{Conditional Generator Network Architecture}
\vspace{-2.0mm}
The generator network aims to generate super-resolved  representations for small objects to improve detection accuracy. To achieve this purpose, we design the generator as a deep residual learning network that augments the representations of small objects to super-resolved ones by introducing more fine-grained details absent from the small objects through residual learning.

As shown in Figure~\ref{fig:subnetwork}, the generator takes the feature from the bottom convolutional layer as the input that preserves many low-level details and is informative for feature super-resolution. The resulting feature is first passed into the $3\times3$ convolution filters followed by the $1\times1$ convolution filters to increase the feature dimension to be the same as that of ``Conv5". Then, $B$ residual blocks with the identical layout consisting of two $3\times3$ convolutional filters followed by batch-normalization layer and ReLU activation layer are introduced to learn the residual  representation between the large and the small objects, as a generative model. The learned residual representation is then used to enhance the feature pooled from ``Conv5" for the small object proposal through RoI pooling~\cite{girshick2015fast} by element-wise sum operation, producing super-resolved representation. 

\vspace{-1.3mm}
\subsection{Discriminator Network Architecture}
\vspace{-2.0mm}
\label{sec:g}
As shown in Figure~\ref{fig:subnetwork}, the discriminator network is trained to not only differentiate between the generated super-resolved feature for the small object and the original one from the real large object, but also justify the detection accuracy benefiting from the generated super-resolved feature. Taking the generated super-resolved representation as input, the discriminator passes it into two branches, \textit{i.e.}, the adversarial branch and the perception branch. The adversarial branch consists of two fully-connected layers followed by a sibling output layer with the sigmoid activation, which produces an adversarial loss. The perception branch consists of two fully-connected layers followed by two sibling output layers, which produces a perceptual loss to justify the detection performance contributing to the super-resolved representation. The output units number of the first two fully-connected layers for both branches are $4096$ and $1024$ respectively.

Given the adversarial loss $L_{dis\_a}$ and the perceptual loss $L_{dis_p}$, a final loss function $L_{dis}$ can be produced as weighted sum of both individual loss components. Given weighting parameters $w_1$ and $w_2$, we define $L_{dis}=w_1\times L_{dis\_a} + w_2\times L_{dis_p}$ to encourage the generator network to generate super-resolved  representation with high detection accuracy. Here we set both $w_1$ and $w_2$ to be one. 
\vspace{-4mm}
\paragraph{Adversarial Loss}
Denote $D_{\Theta_{{a}}}$ as the adversarial branch of the discriminator network with parameters $\Theta_{{a}}$. Taking the generated  representation $G_{\Theta_{g}}(F_s)$ for each object proposal as input, this branch outputs the estimated probability of the input  representation belonging to a real large object, denoted as $D_{\Theta_{a}}(G_{\Theta_{g}}(F_s))$. By trying to fool the discriminator network with the generated  representation, an adversarial loss is introduced to encourage the generator network to produce the super-resolved  representation for the small object similar as that of the large object. The adversarial loss $L_{dis\_a}$ is defined as
\vspace{-2.5mm}
\begin{equation}\label{eqn:loss}
	\vspace{-7mm}
	{L_{dis\_a}} = -\log D_{\Theta_{a}}(G_{\Theta_{g}}(F_s)).
\end{equation}
\vspace{-5mm}
\paragraph{Perceptual Loss}
Taking the super-resolved representation for each proposal as input, the perception branch outputs the category-level confidences $p=(p_0,...,p_k)$ for $K+1$ categories and the bounding-box regression offsets, $r_{k}=(r_{x}^k, r_{y}^k, r_{w}^k, r_{h}^k)$ for each of the $K$ object classes, indexed by $k$. Following the parameterization scheme in~\cite{girshick2013rich}, $r_{k}$ specifies a scale-invariant translation and log-space height/width shift relative to an object proposal. Each training proposal is labeled with a ground-truth class $g$ and a ground-truth bounding-box regression target $r^{*}$. The following multi-task loss $L_{dis\_p}$ is computed to justify the detection accuracy benefiting from the generated super-resolved features for each object proposal:
\vspace{-2.5mm}
\begin{equation}\label{eqn:loss}
	\vspace{-2.5mm}
	{L_{dis\_p}}=L_{cls}(p,g)+\mathbf{1}[g\geq 1]L_{loc}(r_{g},r^{*}),
\end{equation}
where $L_{cls}$ and $L_{loc}$ are the losses for the classification and the bounding-box regression, respectively. In particular, $L_{cls}(p,g) = -\log p_{g}$ is log loss for the ground truth class $g$ and $L_{loc}$ is a smooth $L_1$ loss proposed in~\cite{girshick2015fast}. For background proposals (\emph{i.e. }$g=0$), the $L_{loc}$ is ignored.


\vspace{-2mm}

\section{Experiments}
\vspace{-1mm}

\subsection{Datasets and Evaluation Metrics}
\vspace{-2mm}
\subsubsection{Traffic-sign Detection Datasets}
\vspace{-2mm}
The Tsinghua-Tencent 100K~\cite{zhu2016traffic} is a large traffic-sign benchmark, which contains 30,000 traffic-sign instances. The images are of resolution 2,048$\times$2,048. Following~\cite{zhu2016traffic}, we ignore the classes whose instances are less than 100 and  have 45 classes left. The performance is evaluated using the same detection metrics as for the Microsoft COCO benchmark. We report the detection performance on difference sizes of objects, including small objects (area $<$ $32\times32$ pixels), medium objects ($32\times32$ $<$ area $<$ $96\times96$) and large objects (area $>$ $96\times96$). The numbers of instances corresponding to the three kinds of division are $3270$, $3829$ and $599$, respectively. This evaluation scheme helps us understand the ability of a detector on  objects of different sizes.

%

\vspace{-6mm}
\subsubsection{Pedestrian Detection Datasets}
\vspace{-2mm}
The Caltech benchmark~\cite{dollar2012pedestrian} is  the most popular pedestrian detection dataset. About 250,000 frames with a total of 350,000 bounding boxes and 2,300 unique pedestrians are annotated. We use dense sampling of the training data (every 4th frame) as adopted in~\cite{compact,nam2014local}. Following the conventional evaluation setting~\cite{dollar2012pedestrian}, the performance is evaluated on pedestrians over 50 pixels tall with no or partial occlusion, which are often of very small sizes. The evaluation metric is log-average Miss Rate on False Positive Per Image (FPPI) in $[10^{-2},10^0]$ (denoted as $MR$ following~\cite{zhang2016far}). 


\subsection{Implementation Details}
\vspace{-1.5mm}
For traffic sign detection, we use the pretrained VGG-CNN-M-1024 model~\cite{chatfield2014return} as adopted in~\cite{liu2016deep} to initialize our network. For pedestrian detection, we use the pretrained VGG-16 model~\cite{simonyan2014very} as adopted in~\cite{zhang2016faster}.  For the generator and the discriminator network, the parameters of newly added convolutional layers and fully connected layers are initialized with ``Xavier''~\cite{glorot2010understanding}.  We resize the image to 1600 pixels and 960 pixels on the shortest side as input for traffic sign detection and pedestrian detection respectively. Following~\cite{he2015deep}, we perform down-sampling directly by convolutional layers with a stride of 2. The implementation is based on the publicly available Fast R-CNN framework~\cite{girshick2015fast} built on the Caffe platform~\cite{jia2014caffe}.

The whole network is trained with Stochastic Gradient Descent (SGD) with momentum of 0.9, and weight decay of 0.0005 on a single NVIDIA GeForce GTX TITAN X GPU with 12GB memory. For training the generator network, each SGD mini-batch contains 128 selected object proposals from each training image. Following~\cite{girshick2015fast}, in each mini-batch, 25\% of object proposals are foreground that overlap with a ground truth bounding box with at least 0.5 IoU, and the rest are background. For training the discriminator network, each SGD mini-batch contains 32 selected foreground object proposals from four training images. The number of residual blocks in the generator network $B$ is set as $6$. 
For the Tsinghua-Tencent 100K~\cite{zhu2016traffic} benchmark, we train a Region Proposal Network (RPN) as proposed in~\cite{ren2015faster} to generate object proposals on the training and testing images. For the Caltech benchmark~\cite{dollar2012pedestrian}, we utilize the ACF pedestrian detector~\cite{dollar2014fast} trained on the Caltech training set for object proposals generation. For testing, on average, the Perceptual GAN processes one image within $0.6$ second (excluding object proposal time).



\vspace{-1mm}
\subsection{Performance Comparison}
\vspace{-1mm}
\subsubsection{Traffic-sign Detection}
\begin{figure*}[h]
	\centering
	\includegraphics[scale=0.5]{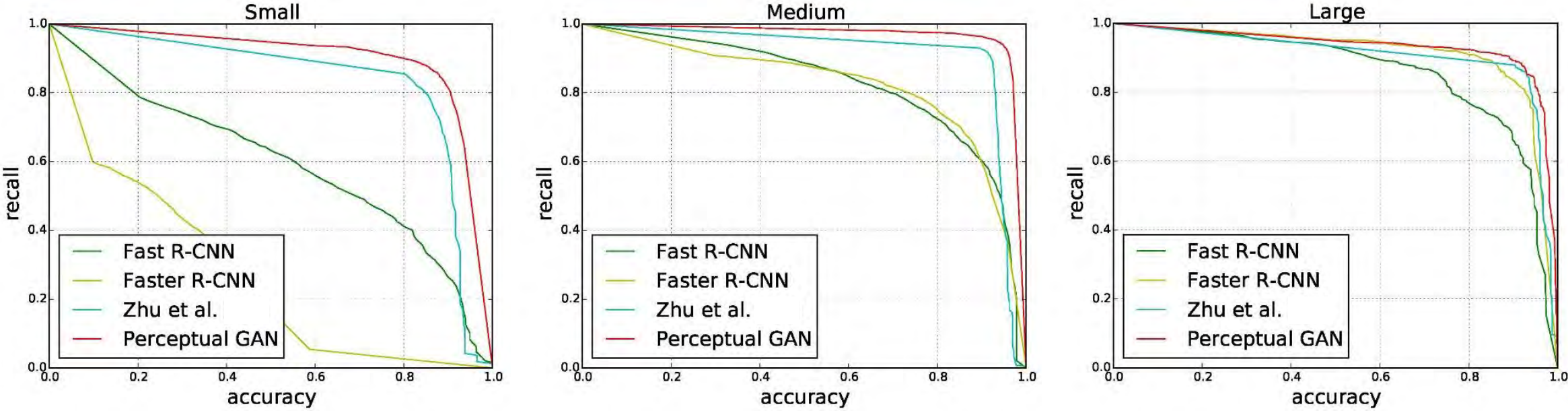}
	\caption{Comparisons of overall detection performance on Tsinghua-Tencent 100K, for small, medium and large traffic signs.}
	\label{fig:results}
	\vspace{-2mm}
\end{figure*}

\vspace{-2mm}
Table~\ref{tab:results_average} provides the comparison of our approach with other state-of-the-arts in terms of average recall and accuracy on traffic-sign detection. It can be observed that the proposed Perceptual GAN outperforms the previous state-of-the-art method of Zhu \emph{et al.}~\cite{zhu2016traffic} in terms of average recall and accuracy: $89\%$ and $84\%$ vs $87\%$ and $82\%$, $96\%$ and $91\%$ vs $94\%$ and $91\%$, $89\%$ and $91\%$ vs $88\%$ and $91\%$ on three subsets of different object sizes. Specifically, our approach makes a large improvement, \emph{i.e.}, $2\%$ and $2\%$ in average recall and accuracy on the small-size subset, demonstrating its superiority in accurately detecting small objects. Table~\ref{tab:results_class} shows the comparisons of recall and accuracy for each category. Our approach achieves the best performance in most categories such as ``p3" and ``pm55" in which small instances are most common. More comparisons of accuracy-recall curves in terms of different object sizes are provided in Figure~\ref{fig:results}, which can further demonstrate the effectiveness of the proposed generative adversarial learning strategy.


\begin{table}\setlength{\tabcolsep}{3pt}
	\centering\scriptsize\tabcolsep=0.2cm
	\caption{Comparisons of detection performance for different sizes of traffic signs on Tsinghua-Tencent 100K. (R): Recall, (A): Accuracy. (In $\%$)}\label{tab:results_average}
	\renewcommand{\arraystretch}{1.3}
	\begin{tabular}{c | c | c | c}
		\toprule
		{\bf Object size} & {\bf Small} & {\bf Medium} & {\bf Large}\\ \hline
		Fast R-CNN~\cite{girshick2015fast} (R) & 46 & 71 & 77 \\
		Fast R-CNN~\cite{girshick2015fast} (A) & 74 & 82 & 80 \\ \hline
		Faster R-CNN~\cite{ren2015faster} (R) & 50 & 84 & 91 \\
		Faster R-CNN~\cite{ren2015faster} (A) & 24 & 66 & 81 \\ \hline
		Zhu \emph{et al.}~\cite{zhu2016traffic} (R) & 87 & 94 & 88 \\
		Zhu \emph{et al.}~\cite{zhu2016traffic} (A) & 82 & 91 & 91 \\ \hline
		\textbf{Ours (R)} & \textbf{89} & \textbf{96} & \textbf{89} \\
		\textbf{Ours (A)} & \textbf{84} & \textbf{91} & \textbf{91} \\ 				
		\bottomrule
	\end{tabular}%
	\vspace{-4mm}
\end{table}%

\begin{figure*}[h]
	\centering
	\includegraphics[scale=0.37]{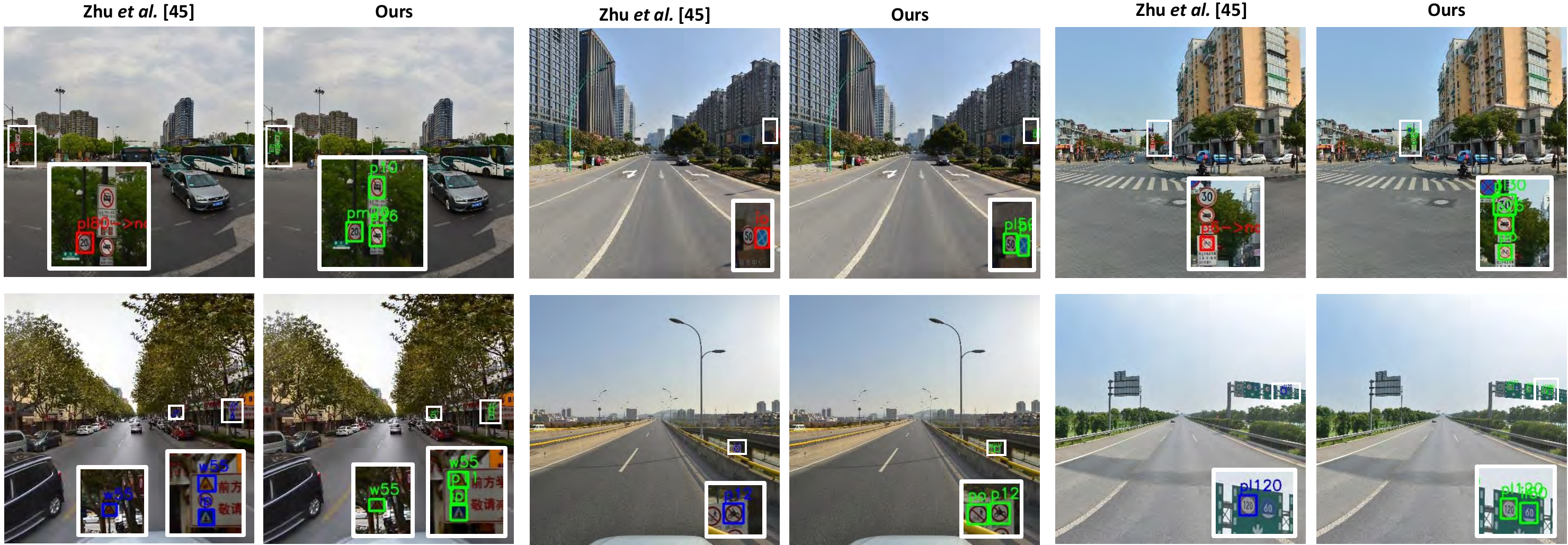}
	\caption{Detection results of Zhu \emph{et al.}~\cite{zhu2016traffic} and the proposed method on Tsinghua-Tencent 100K. The green, red, and blue rectangle denote the true positive, false positive and false negative respectively. The proposed Perceptual GAN can successfully detect most small-size traffic signs which the method of Zhu \emph{et al.}~\cite{zhu2016traffic} has missed or detected incorrectly. Best viewed in color. }
	\label{fig:visualization}
	\vspace{-4mm}
\end{figure*}

Several examples of the detection results for small objects are visualized in Figure~\ref{fig:visualization}. We compare our visual results with those from Zhu \emph{et al.}~\cite{zhu2016traffic}. Note that Zhu \emph{et al.}~\cite{zhu2016traffic} take the original image of resolution $2,048 \times 2,048$ as input, which may cause heavy time consumption for training and testing. In contrast, the Perceptual GAN only takes image of resolution $1600 \times 1600$ as input. In addition, no data augmentation as adopted by Zhu \emph{et al.}~\cite{zhu2016traffic} has been applied. As shown in Figure~\ref{fig:visualization}, generally, our method can accurately classify and localize most objects in small scales, while Zhu \emph{et al.}~\cite{zhu2016traffic} fails to localize some instances due to serious small-scale problem. 

\vspace{-4mm}
\subsubsection{Pedestrian Detection}
\begin{figure}[t]
	\centering
	\includegraphics[scale=0.45]{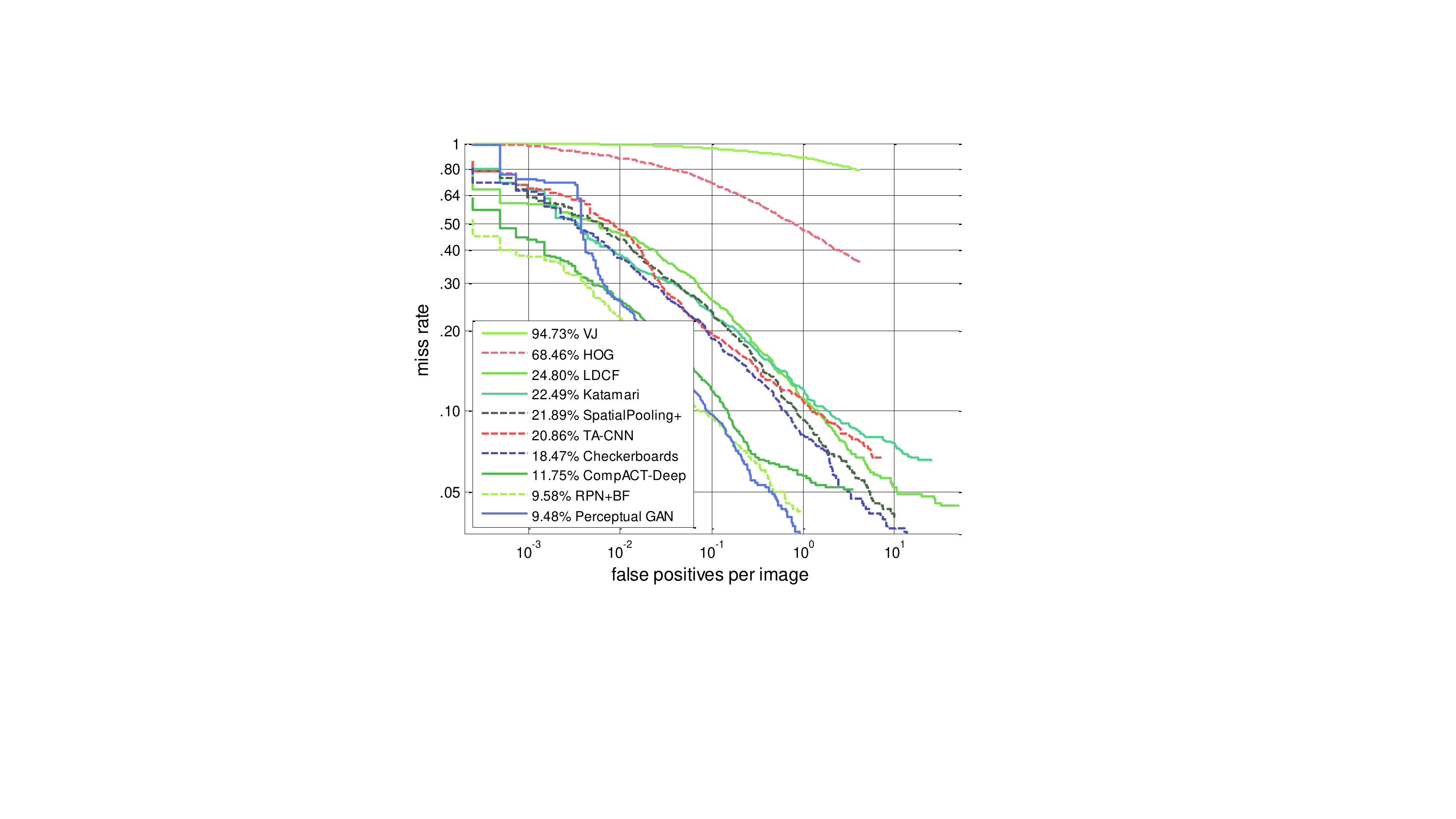}
	\caption{Comparisons of detection performance with the state-of-the-arts on the Caltech benchmark.}
	\label{fig:result_Caltech}
	\vspace{-1em}
\end{figure}

\vspace{-2mm}
Since the pedestrian instances on the Caltech benchmark~\cite{dollar2012pedestrian} are often of small scales, the overall performance on it can be used to evaluate the capability of a method in detecting small objects. We compare the result of Perceptual GAN with all the existing methods that achieved best performance on the Caltech testing set, including VJ~\cite{viola2004robust}, HOG~\cite{dalal2005histograms}, LDCF~\cite{nam2014local}, Katamari~\cite{benenson2014ten}, SpatialPooling+~\cite{paisitkriangkrai2014strengthening}, TA-CNN~\cite{ta_cnn}, Checkerboards~\cite{zhang2015filtered}, CompACT-Deep~\cite{compact} and RPN+BF~\cite{zhang2016faster}. As shown in Figure~\ref{fig:result_Caltech}, the proposed method outperforms all the previous methods and achieves the lowest log-average miss rate of $9.48\%$, validating its superiority in detecting small objects.

\begin{table*} [htbp]\setlength{\tabcolsep}{3pt}
	\centering\scriptsize\tabcolsep=0.15cm
	{\caption{Comparisons of detection performance for each class on Tsinghua-Tencent 100K. (R): Recall, (A): Accuracy. (In $\%$)} \label{tab:results_class}	
		\begin{tabular}{c ccccccccccccccc}
			\toprule[1.0pt]
			\textbf{Class} & \textbf{i2} & \textbf{i4} & \textbf{i5} & \textbf{il100} & \textbf{il60} & \textbf{il80} & \textbf{io} & \textbf{ip} & \textbf{p10} & \textbf{p11} & \textbf{p12} & \textbf{p19} & \textbf{p23} & \textbf{p26} & \textbf{p27} \\
			\midrule
			Fast R-CNN~\cite{girshick2015fast} (R) & 51 & 74 & 84 & 44 & 61 & 10 & 70 & 73 & 54 & 71 & 21 & 42 & 65 & 63 & 36 \\
			Fast R-CNN~\cite{girshick2015fast} (A) & 82 & 86 & 85 & 85 & 70 & 91 & 75 & 80 & 72 & 73 & 47 & 48 & 79 & 74 & 100  \\
			Faster R-CNN~\cite{ren2015faster} (R) & 60 & 76 & 80 & 74 & 89 & 77 & 72 & 64 & 62 & 61 & 53 & 73 & 75 & 78 & 81 \\
			Faster R-CNN~\cite{ren2015faster} (A) & 44 & 46 & 45 & 41 & 57 & 62 & 41 & 39 & 45 & 38 & 60 & 59 & 65 & 50 & 79 \\
			Zhu \emph{et al.}~\cite{zhu2016traffic} (R) & 82 & 94 & 95 & 97 & 91 & 94 & 89 & 92 & 95 & 91 & 89 & 94 & 94 & 93 & 96 \\
			Zhu \emph{et al.}~\cite{zhu2016traffic} (A) & 72 & 83 & 92 & 100 & 91 & 93 & 76 & 87 & 78 & 89 & 88 & 53 & 87 & 82 & 78 \\
			\textbf{Ours (R)} & \textbf{84} & \textbf{95} & \textbf{95} & \textbf{95} & \textbf{92} & \textbf{95} & \textbf{92} & \textbf{91} & \textbf{89} & \textbf{96} & \textbf{97} & \textbf{97} & \textbf{95} & \textbf{94} & \textbf{98} \\
			\textbf{Ours (A)} & \textbf{85} & \textbf{92} & \textbf{94} & \textbf{97} & \textbf{95} & \textbf{83} & \textbf{79} & \textbf{90} & \textbf{84} & \textbf{85} & \textbf{88} & \textbf{84} & \textbf{92} & \textbf{83} & \textbf{98} \\
			\midrule[0.8pt]			
			\midrule[0.8pt]
			\textbf{Class} & \textbf{p3} & \textbf{p5} & \textbf{p6} & \textbf{pg} & \textbf{ph4} & \textbf{ph4.5} & \textbf{ph5} & \textbf{pl100} & \textbf{pl120} & \textbf{pl20} & \textbf{pl30} & \textbf{pl40} & \textbf{pl5} & \textbf{pl50} & \textbf{pl60} \\
			\midrule
			Fast R-CNN~\cite{girshick2015fast} (R) & 50  & 78 & 8  & 88 & 32 & 77 & 18 & 68 & 39 & 14 & 18 & 58 & 69 & 34 & 41 \\
			Fast R-CNN~\cite{girshick2015fast} (A) & 85 & 87 & 100  & 86 & 92 & 82 & 88 & 86 & 92 & 89 & 59 & 78 & 88 & 65 & 73 \\
			Faster R-CNN~\cite{ren2015faster} (R) & 55 & 82 & 54 & 84 & 57 & 80 & 46 & 86 & 77 & 46 & 61 & 68 & 69 & 62 & 65 \\
			Faster R-CNN~\cite{ren2015faster} (A) & 48 & 57 & 75 & 80 & 68 & 58 & 51 & 68 & 67 & 51 & 43 & 52 & 53 & 39 & 53 \\
			Zhu \emph{et al.}~\cite{zhu2016traffic} (R) & 91 & 95 & 87 & 91 & 82 & 88 & 82 & 98 & 98 & 96 & 94 & 96 & 94 & 94 & 93 \\
			Zhu \emph{et al.}~\cite{zhu2016traffic} (A) & 80 & 89 & 87 & 93 & 94 & 88 & 89 & 97 & 100 & 90 & 90 & 89 & 84 & 87 & 93 \\
			\textbf{Ours (R)} & \textbf{93} & \textbf{96} & \textbf{100}  & \textbf{93} & \textbf{78} & \textbf{88} & \textbf{85} & \textbf{96} & \textbf{98} & \textbf{96} & \textbf{93} & \textbf{96} & \textbf{92} & \textbf{96} & \textbf{91} \\
			\textbf{Ours (A)} & \textbf{92} & \textbf{90} & \textbf{83} & \textbf{93} & \textbf{97} & \textbf{68} & \textbf{69} & \textbf{97} & \textbf{98} & \textbf{92} & \textbf{91} & \textbf{90} & \textbf{86} & \textbf{87} & \textbf{92} \\
			\midrule[0.8pt]	
			\midrule[0.8pt]
			\textbf{Class} & \textbf{pl70} & \textbf{pl80} & \textbf{pm20} & \textbf{pm30} & \textbf{pm55} & \textbf{pn} & \textbf{pne} & \textbf{po} & \textbf{pr40} & \textbf{w13} & \textbf{w32} & \textbf{w55} & \textbf{w57} & \textbf{w59} & \textbf{wo} \\
			\midrule
			Fast R-CNN~\cite{girshick2015fast} (R) & 2  & 34 & 43 & 19 & 58 & 87 & 90 & 46 & 95 & 32 & 41 & 43 & 73 & 74 & 16 \\
			Fast R-CNN~\cite{girshick2015fast} (A) & 100  & 84 & 70  & 67 & 76 & 85 & 87 & 66 & 78 & 40  & 100  & 57 & 66 & 64 & 55 \\
			Faster R-CNN~\cite{ren2015faster} (R) & 68 & 68 & 63 & 63 & 79 & 77 & 83 & 63 & 98 & 71 & 59 & 63 & 79 & 78 & 50 \\
			Faster R-CNN~\cite{ren2015faster} (A) & 61 & 52 & 61 & 67 & 61 & 37 & 47 & 37 & 75 & 33 & 54 & 39 & 48 & 39 & 37 \\
			Zhu \emph{et al.}~\cite{zhu2016traffic} (R) & 93 & 95 & 88 & 91 & 95 & 91 & 93 & 67 & 98 & 65 & 71 & 72 & 79 & 82 & 45 \\
			Zhu \emph{et al.}~\cite{zhu2016traffic} (A) & 95 & 94 & 91 & 81 & 60 & 92 & 93 & 84 & 76 & 65 & 89 & 86 & 95 & 75 & 52 \\
			\textbf{Ours (R)} & \textbf{91} & \textbf{99} & \textbf{88} & \textbf{94} & \textbf{100}  & \textbf{96} & \textbf{97} & \textbf{83} & \textbf{97} & \textbf{94} & \textbf{85} & \textbf{95} & \textbf{94} & \textbf{95} & \textbf{53} \\
			\textbf{Ours (A)} & \textbf{97} & \textbf{86} & \textbf{90} & \textbf{77} & \textbf{81} & \textbf{89} & \textbf{93} & \textbf{78} & \textbf{92} & \textbf{66} & \textbf{83} & \textbf{88} & \textbf{93} & \textbf{71} & \textbf{54} \\											
			\bottomrule[1.0pt]					             			              			
		\end{tabular}
		\vspace{-4mm}
	}	
\end{table*}

\vspace{-1mm}
\subsection{Ablation Studies}
\vspace{-1mm}
We investigate the effectiveness of different components of Perceptual GAN. All experiments are performed on the Tsinghua-Tencent 100K~\cite{zhu2016traffic} dataset. The performance achieved by different variants of Perceptual GAN and parameter settings on small objects and all the objects of different sizes are reported in the following.

\begin{table}\setlength{\tabcolsep}{3.5pt}
	\centering\scriptsize\tabcolsep=0.3cm
	\caption{Comparisons of detection performance with several variants of Perceptual GAN on Tsinghua-Tencent 100K. (R): Recall, (A): Accuracy. (In $\%$)}\label{tab:variants}
	\renewcommand{\arraystretch}{1.3}
	\begin{tabular}{c | c | c}
		\toprule
		{\bf Object size} & {\bf Small} & {\bf All} \\ \hline	
		Skip Pooling (R) & 76 & 87 \\
		Skip Pooling (A) & 82 & 86 \\ \hline
		Large Scale Images (R) & 85 & 92 \\
		Large Scale Images (A) & 81 & 86 \\	\hline	
		Multi-scale Input (R) & 89 & 93 \\
		Multi-scale Input (A) & 77 & 83 \\ \hline
		\textbf{Ours (R)} & \textbf{89} & \textbf{93} \\
		\textbf{Ours (A)} & \textbf{84} & \textbf{88} \\
		\bottomrule
	\end{tabular}%
	\vspace{-6mm}
\end{table}%

\begin{figure}[t]
	\centering
	\includegraphics[scale=0.43]{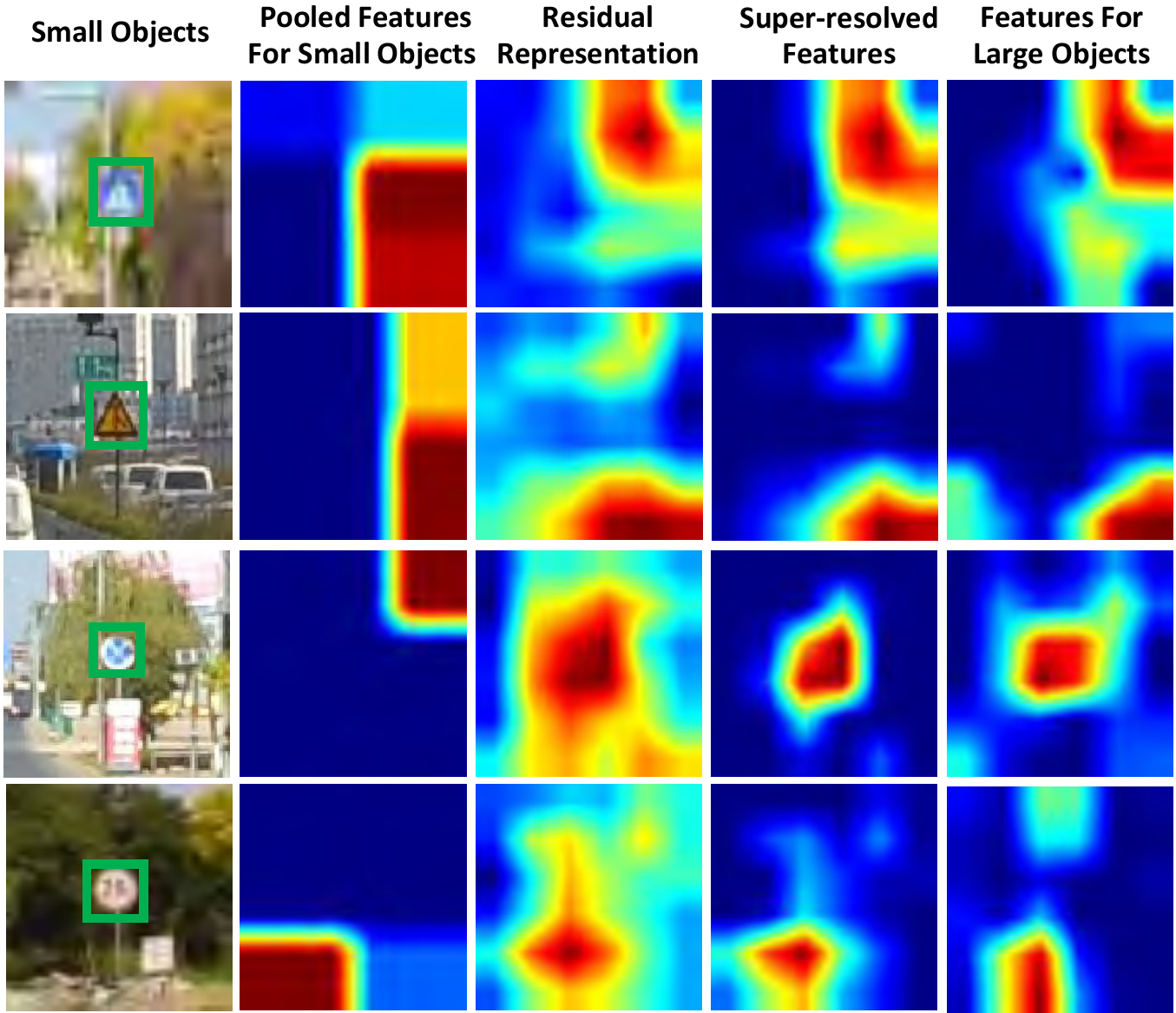}
	\caption{Visualization of the super-resolved features.}
	\label{fig:Super-resolved}
	\vspace{-6mm}
\end{figure}

\vspace{-4mm}
\subsubsection{The Effectiveness of Super-resolved Features by Generator}
\vspace{-2mm}
To verify the superiority of the generated super-resolved  representation in detecting small objects, we compare our method with several other feature enhancement solutions, including combining low-level features, improving the image resolution by simply increasing the input scales, taking images with multi-scales as input. All these methods are implemented based on the base convolutional layers and the perceptual branch with end-to-end training. As shown in Table~\ref{tab:variants}, ``\textbf{Skip Pooling}" indicates the model trained by combining low-level features through skip pooling as proposed in ~\cite{bell2015inside}. Our Perceptual GAN outperforms this approach by $13\%$ and $2\%$ in average recall and accuracy on small-size objects respectively, which validates that our method can effectively incorporate fine-grained details from low-level layers to improve small object detection. ``\textbf{Large Scale Images}" represents the model trained with images of higher resolution by simply increasing the scale of input images to $2048\times2048$. ``\textbf{Multi-scale Input}" indicates the model trained with input images with multi-scale settings ($s\in{1120,1340,1600,1920,2300}$) as adopted in~\cite{girshick2015fast}. One can observe that our Perceptual GAN outperforms both approaches in performance on small objects. This shows that our method is more effective in boosting small object detection than simply increasing the input image scale or using multi-scale settings. 

We further visualize some of the generated super-resolved features, as shown in Figure~\ref{fig:Super-resolved}. The second and the last column show the original features pooled from the top convolutional layer for proposals of small objects and large objects respectively. The learned residual representation and the generated super-resolved features by the generator for small objects are shown in the third and the fourth column respectively. One can observe that the generator successfully learns to transfer the poor representations of small objects to super-resolved ones similar to those of large objects, validating the effectiveness of the Perceptual GAN. 

\vspace{-1mm}
\subsubsection{The Effectiveness of Adversarial Training}
\vspace{-2mm}

The proposed Perceptual GAN trains the generator and the discriminator through alternative optimization. To demonstrate the necessity of adversarial training, we report the performance of our model with or without alternative optimization during training stage in Table~\ref{tab:Recursive}. ``Ours\_Baseline" indicates the model of training the proposed detection pipeline with the generator network end-to-end without any alternative optimization step. ``Ours\_Alt" indicates the model of alternatively training the generator and the discriminator. By comparing ``Ours\_Alt" with ``Ours\_Baseline", one can observe that considerable improvements in the recall and accuracy on small-size object detection can be obtained when using alternative optimization. This shows that Perceptual GAN can improve its performance in detecting small objects by recursively improving the ability of the generator and the discriminator through adversarial training. 

\begin{table}\setlength{\tabcolsep}{3.5pt}
	\centering\scriptsize\tabcolsep=0.3cm
	\caption{Comparisons of detection performance by Perceptual GAN with or without alternative optimization on Tsinghua-Tencent 100K. (R): Recall, (A): Accuracy. (In $\%$)}\label{tab:Recursive}
	\renewcommand{\arraystretch}{1.3}
	\begin{tabular}{c | c | c}
		\toprule
		{\bf Object size} & {\bf Small} & {\bf All} \\ \hline	
		Ours\_Baseline (R) & 80 & 89 \\
		Ours\_Baseline (A) & 80 & 85 \\ \hline
		\textbf{Ours\_Alt (R)} & \textbf{89} & \textbf{93} \\
		\textbf{Ours\_Alt (A)} & \textbf{84} & \textbf{88} \\
		\bottomrule
	\end{tabular}%
	\vspace{-1mm}
\end{table}%

\vspace{-4mm}
\subsubsection{Different Lower Layers for Learning Generator}
\vspace{-2mm}
The proposed generator learns fine-grained details of small objects from  representations of lower-level layers. In particular, we employ the features from ``Conv1" as the inputs for learning the generator. To validate the effectiveness of this setting, we conduct additional experiments using features from ``Conv2" and ``Conv3" for learning the generator, respectively. As shown in Table~\ref{tab:Conv}, we can observe that performance consistently decreases by employing the  representations from higher layers. The reason is that lower layers can capture more details of small objects. Therefore, using low-level features from ``Conv1" for learning the generator gives the best performance.

\begin{table}\setlength{\tabcolsep}{3.5pt}
	\centering\scriptsize\tabcolsep=0.3cm
	\caption{Comparisons of detection performance for introducing fine-grained details from different lower-level layers on Tsinghua-Tencent 100K. (R): Recall, (A): Accuracy. (In $\%$)}
	\label{tab:Conv}
	\renewcommand{\arraystretch}{1.3}
	\begin{tabular}{c | c | c}
		\toprule
		{\bf Object size} & {\bf Small} & {\bf All} \\ \hline	
		Ours\_Conv3 (R) & 74 & 86 \\
		Ours\_Conv3 (A) & 78 & 85 \\ \hline
		Ours\_Conv2 (R) & 87 & 92 \\
		Ours\_Conv2 (A) & 80 & 86 \\ \hline
		\textbf{Ours\_Conv1 (R)} & \textbf{89} & \textbf{93} \\
		\textbf{Ours\_Conv1 (A)} & \textbf{84} & \textbf{88} \\
		\bottomrule
	\end{tabular}%
	\vspace{-4mm}
\end{table}%

\vspace{-2mm}
\subsection{Discussion on General Small Object Detection}
\vspace{-2mm}
To evaluate the generalization capability of the proposed generator on more general and diverse object categories, we train the proposed detection pipeline with the generator network end-to-end on the union of the trainval set of PASCAL VOC 2007 and VOC 2012~\cite{everingham2010the}, and evaluate it on the test set of VOC 2007 on the most challenging classes (\emph{i.e.}, boat, bottle, chair and plant) in which small instances are most common. Our method achieves $69.4\%$, $60.2\%$, $57.9\%$ and $41.8\%$ in Average Precision (AP) for boat, bottle, chair, and plant, respectively. It significantly outperforms those of the Fast R-CNN~\cite{girshick2015fast} baseline, \emph{i.e.}, $59.4\%$, $38.3\%$, $42.8\%$ and $31.8\%$, well demonstrating the generalization capability of the proposed generator for general small object detection.

\vspace{-3.3mm}
\section{Conclusion}
\vspace{-2.5mm}
In this paper, we proposed a novel generative adversarial network to address the challenging problem of small object detection. Perceptual GAN generates super-resolved  representations for small objects to boost detection performance by leveraging the repeatedly updated generator network and the discriminator network. The generator learns a residual  representation from the fine-grained details from lower-level layers, and enhances the  representations for small objects to approach those for large objects by trying to fool the discriminator which is trained to well differentiate between both representations. Competition in the alternative optimization of both networks encourages the Perceptual GAN to generate super-resolved large-object like  representations for small objects, thus improving detection performance. Extensive experiments have demonstrated the superiority of the proposed Perceptual GAN in detecting small objects.

\vspace{-5mm}
\section*{Acknowledgement}
\vspace{-3mm}
This work was partially supported by China Scholarship Council (Grant No. 201506030045). The work of Jiashi Feng was partially supported by National University of Singapore startup grant R-263-000-C08-133 and Ministry of Education of Singapore AcRF Tier One grant R-263-000-C21-112.

{\footnotesize
\bibliographystyle{ieee}
\bibliography{egbib}
}

\end{document}